\documentclass[conference]{IEEEtran}
\IEEEoverridecommandlockouts

\usepackage{cite}
\usepackage{amsmath,amssymb,amsfonts}
\usepackage{algorithmic}
\usepackage{graphicx}
\usepackage{textcomp}
\usepackage{xcolor}
\def\BibTeX{{\rm B\kern-.05em{\sc i\kern-.025em b}\kern-.08em
    T\kern-.1667em\lower.7ex\hbox{E}\kern-.125emX}}
\begin{document}

\title{When Small Variations Become Big Failures: Reliability Challenges in Compute-in-Memory Neural Accelerators\\
}

\author{
    \textbf{
        Yifan Qin\textsuperscript{1} \ \ \ \ 
        Jiahao Zheng\textsuperscript{1} \ \ \ \ 
        Zheyu Yan\textsuperscript{1} \ \ \ \ 
        Wujie Wen\textsuperscript{2} \ \ \ \ 
        Xiaobo Sharon Hu\textsuperscript{1} \ \ \ \ 
        Yiyu Shi\textsuperscript{1}
    }\\
    \IEEEauthorblockA{\textsuperscript{1}University of Notre Dame, \textsuperscript{2}NC State University, \{yqin3, shu, yshi4\}@nd.edu}
    \vspace{-1cm}
}

\maketitle

\begin{abstract}
Compute-in-memory (CiM) architectures promise significant improvements in energy efficiency and throughput for deep neural network acceleration by alleviating the von Neumann bottleneck. However, their reliance on emerging non-volatile memory devices introduces device-level non-idealities—such as write variability, conductance drift, and stochastic noise—that fundamentally challenge reliability, predictability, and safety, especially in safety-critical applications. This talk examines the reliability limits of CiM-based neural accelerators and presents a series of techniques that bridge device physics, architecture, and learning algorithms to address these challenges. We first demonstrate that even small device variations can lead to disproportionately large accuracy degradation and catastrophic failures in safety-critical inference workloads, revealing a critical gap between average-case evaluations and worst-case behavior. Building on this insight, we introduce SWIM, a selective write-verify mechanism that strategically applies verification only where it is most impactful, significantly improving reliability while maintaining CiM’s efficiency advantages. Finally, we explore a learning-centric solution that improves realistic worst-case performance by training neural networks with right-censored Gaussian noise, aligning training assumptions with hardware-induced variability and enabling robust deployment without excessive hardware overhead.
Together, these works highlight the necessity of cross-layer co-design for CiM accelerators and provide a principled path toward dependable, efficient neural inference on emerging memory technologies—paving the way for their adoption in safety- and reliability-critical systems.
\end{abstract}

\begin{IEEEkeywords}
compute-in-memory, variation, reliability
\end{IEEEkeywords}

\vspace{-0.3cm}
\section{Introduction}
Emerging non-volatile compute-in-memory (NVCiM) deep neural network accelerators~\cite{qin2024tsb,yan2023improving,diware2023mapping} promise substantial energy efficiency and throughput benefits by reducing costly data movement, yet their deployment is fundamentally constrained by reliability. NVM device intrinsic non-idealities~\cite{qin2020design}, including read noise, conductance drift, write variations, introduce weights noise to deployed neural network models and decrease their performance, especially for safety-critical scenarios. A key challenge is that CiM robustness cannot be assessed by average-case accuracy alone: rare but possible device variations may trigger massive accuracy loss and even catastrophic failures, creating a gap between conventional evaluation and real deployment requirements. This paper summarizes three complementary works that collectively diagnose and mitigate such failures through hardware/software co-design:
\begin{itemize}
    \item \textbf{Worst-case characterization:} we show that even small device variations can cause unexpectedly large degradation~\cite{yan2022computing} under realistic settings, highlighting the necessity of tail-aware evaluation beyond Monte Carlo averages.
    \item \textbf{Hardware design:} we introduce selective write-verify (SWIM)~\cite{yan2022swim} that applies verification only to the most impactful weight devices under a budget, improving reliability while preserving operation efficiency.
    \item \textbf{Software design:} We propose a realistic worst-case metric for evaluation and improve it by training with right-censored Gaussian noise~\cite{yan2023improving} without excessive overhead.
\end{itemize}
The remainder of this paper is organized as follows: Section~\ref{sec:worstcase} presents worst-case behaviors under NVM device variations, and Section~\ref{sec:solutions} introduces cross-layer solutions based on selective write-verify and right-censored Gaussian noise training.

\vspace{-0.2cm}
\section{Worst-Case Failures under Variations}\label{sec:worstcase}
Prior studies~\cite{11087589} of NVCiM accelerators predominantly report \emph{average-case} accuracy under device variations which causes weight bounded but independent Gaussian noise. However, for safety-critical inference, the \emph{worst-case} behavior is often the determining factor: rare but plausible combinations of device variations can dominate risk and lead to catastrophic outcomes. This challenge is exacerbated by the high-dimensional variation space, making it unlikely for na\"ive Monte Carlo simulations to capture the true tail behavior even with a large number of runs.

To quantify this gap, we formulate worst-case reliability evaluation as an optimization problem~\cite{yan2022computing}: given a realistic weight noise $\Delta W$ with write-verify enforced bounds $th_g$, we search for the specific combination of weight noise that minimizes inference performance (equivalently, minimizes accuracy):
\begin{equation}
\begin{gathered}
\min_{\Delta W} \left | \left \{ f\left ( W + \Delta W, x \right ) = t \middle | \left( x, t \right ) \in D\right \}  \right | \\
\text{s.t. }\ \mathcal{L}(\Delta W)\le th_g 
\end{gathered}
\end{equation}
In practical NVCiM settings where write-verify bounds the noise perturbation, our findings reveal a striking amplification effect: although each individual variation is small, the jointly worst-case configuration can drive \emph{drastic} accuracy collapse, with the worst-case error approaching 100\% on representative networks and datasets. In contrast, MC sampling may appear ``converged'' in terms of average statistics (even at 100K runs) yet still misses these tail failures by a wide margin.

Importantly, this worst-case lens also exposes a limitation of existing mitigation strategies that primarily optimize average performance under variations: techniques that are effective in improving mean accuracy can be surprisingly ineffective when directly extended to improve worst-case reliability. These observations motivate a cross-layer solutions in Section~\ref{sec:solutions}: we propose (i) \emph{hardware} selective write-verify (SWIM) to pick the most impactful noise and decrease operation cost, and (ii) \emph{software} right-censored noise training to improve realistic worst-case robustness.

\section{Cross-Layer Solutions}\label{sec:solutions}
Motivated by the disproportionate worst-case failures in Section~\ref{sec:worstcase}, we summarize two complementary mitigation strategies for dependable NVCiM inference. SWIM is a \emph{hardware} approach that selectively applies write-verify operations to the most impactful weight devices under a budget, avoiding the high cost of exhaustive operations. Right-censored noise training is a \emph{software} approach that improves realistic worst-case robustness (e.g., k-th percentile performance, or KPP) by injecting training-time explicit noise patten to network model.

\subsection{SWIM: Selective Write-Verify}
Write-verify is a direct and effective operation to suppress programming variations in NVCiM, but exhaustively operate every device incurs prohibitive write latency/energy and undermines the efficiency benefits of CiM. SWIM addresses this tension by selectively applying write-verify only to a small subset of impactful weight devices under a user-specified budget, achieving a trade-off between reliability and efficiency.

SWIM formulates selective write-verify as a budgeted selection problem. Given a mapped model and a performance target (e.g., an allowed accuracy drop $\Delta Acc$ or a minimum accuracy threshold), SWIM seeks a minimal subset of weights to verify such that inference accuracy under variations satisfies the target, while the overall write overhead is minimized. To quantify overhead, SWIM uses a normalized write cycles where a smaller value indicates fewer operations respect to exhaustive write-verify.
A key observation is that na\"ive heuristics for choosing verified weights are unreliable. For instance, verifying weights by magnitude or layer order often correlates poorly with accuracy degradation under device variations. Instead, SWIM prioritizes weights based on their impact on the objective, using a loss-based sensitivity metric derived from a Taylor-expansion approximation. Intuitively, weight noise that lead larger loss increase when perturbed are more likely to trigger tail failures and therefore should be verified first. SWIM computes these sensitivities efficiently, ranks weights accordingly, and performs write-verify in descending order of sensitivity with a hardware-aware granularity (e.g., per programming group/row), stopping once the accuracy constraint is met.

\subsection{Learning with Right-Censored Noise}
While Section~\ref{sec:worstcase} highlights that true worst-case failures can be catastrophic, such extreme configurations may be exceedingly rare and are difficult to use as a direct engineering objective in practice. We therefore adopt a more realistic tail-robustness metric, the k-th percentile performance (KPP), and present a training method that improves this metric under device intrinsic variations.
\begin{figure}[ht]
\vspace{-0.4cm}
\begin{center}
\centerline{\includegraphics[trim=25 85 40 110, clip, width=0.8\linewidth] {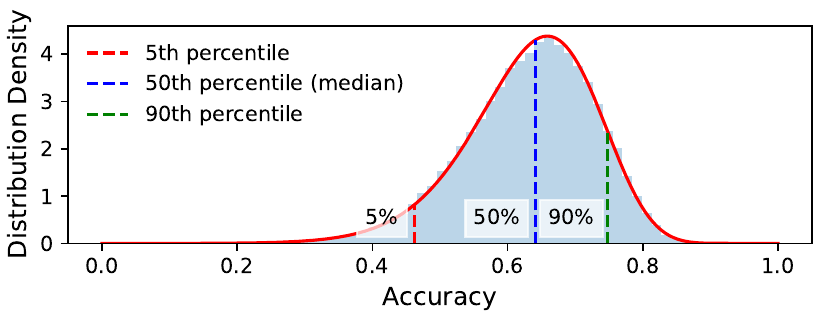}}
\vspace{-0.2cm}
\caption{Illustration of KPP~(in terms of accuracy). Red curve means the accuracy distribution of a DNN in the presence of device variations.}
\label{fig:kthp}
\end{center}
\vspace{-0.6cm}
\end{figure}

Specifically, the KPP is the accuracy threshold such that only the worst $k\%$ variation instances fall below it (Fig.~\ref{fig:kthp}). Unlike the absolute worst-case, KPP is stable and actionable while still capturing tail risk beyond average-case metrics. We focus on small $k$ (e.g., $k{=}1$) as a realistic worst-case proxy for reliability critical deployments. To improve KPP, we propose \emph{TRICE} (\underline{T}raining with \underline{R}Ight-\underline{C}ensored Gaussian Nois\underline{E}), a tail-oriented noise-injection method that trains with right-censored Gaussian noise on weights. This design is motivated by a Taylor-based analysis of loss/gradient under weight perturbations, which indicates that uncensored Gaussian tails can dominate optimization without effectively improving percentile metrics. TRICE is a plug-and-play training method and consistently improves KPP across models and variation strengths without additional hardware overhead.

\section{Conclusion}
This paper summarizes three complementary works toward dependable NVCiM inference: worst-case evaluation exposes catastrophic tail failures under small device variations; SWIM mitigates them via budgeted selective write-verify; and right-censored noise injection improves realistic worst-case robustness by optimizing the k-th percentile performance. These results highlight that reliable NVCiM deployment requires safety-critical metrics and cross-layer co-design over devices, architecture, and learning algorithm.

\bibliographystyle{IEEEtran}
\bibliography{reference}


\end{document}